\documentclass[a4paper,10pt,onecolumn]{article}

\newcommand{\be}{\begin{equation}}
\newcommand{\ee}{\end{equation}}
\newcommand{\beq}{\begin{equation}}
\newcommand{\eeq}{\end{equation}}
\newcommand{\bed}{\begin{displaymath}}
\newcommand{\eed}{\end{displaymath}}
\newcommand{\beqa}{\begin{eqnarray}}
\newcommand{\eeqa}{\end{eqnarray}}
\newcommand{\beqann}{\begin{eqnarray*}}
\newcommand{\eeqann}{\end{eqnarray*}}
\newcommand{\bseq}{\begin{subequation}}
\newcommand{\eseq}{\end{subequation}}

\newcommand{\ba}{\begin{array}}
\newcommand{\ea}{\end{array}}

\newcommand{\negr}[1]{{\bf {#1}}}

\usepackage[dvips]{graphicx}
\usepackage{psfrag}
\usepackage{epsfig}
\usepackage{array}
\usepackage{amssymb}
\usepackage{subeqn}
\usepackage{subfigure}

 \makeatletter
\begin{document}
\date{}
\title{
 \small Proceedings of CK2005, International Workshop on Computational Kinematics \\
 Cassino May 4-6, 2005\\
 paper xxCK2005 \vskip 15mm
 {\Large \bf STRATEGIES FOR THE DESIGN \goodbreak  \goodbreak \vskip 8mm OF A SLIDE-O-CAM TRANSMISSION}}
 \author{D. Chablat$^1$, J. Angeles$^2$ \\
      $^1$Institut de Recherche en Communications et Cybern\'etique de
      Nantes
      \thanks{IRCCyN: UMR n$^\circ$ 6597 CNRS, \'Ecole Centrale de Nantes,
                        Universit\'e de Nantes, \'Ecole des Mines de
                        Nantes}\\
      UMR CNRS n$^\circ$ 6597, 1 rue de la No\"e, 44321 Nantes, France \\
      $^2$Department of Mechanical Engineering \& \\
      Centre for Intelligent Machines, McGill University \\
      817 Sherbrooke Street West, Montreal, Canada H3A 2K6 \\
      Damien.Chablat@irccyn.ec-nantes.fr $\quad$ angeles@cim.mcgill.ca
    }
\maketitle
\thispagestyle{empty}
\subsection*{\centering Abstract}
The optimization of the pressure angle in a cam-follower
transmission is reported in this paper. This transmission is based
on {\em Slide-o-Cam}, a cam mechanism with multiple rollers
mounted on a common translating follower. The design of
Slide-o-Cam, a transmission intended to produce a sliding motion
from a turning drive, or vice versa,  was reported elsewhere. This
transmission provides pure-rolling motion, thereby reducing the
friction of rack-and-pinions and linear drives. The pressure angle
is a suitable performance index for this transmission because it
determines the amount of force transmitted to the load vs.\ that
transmitted to the machine frame. Two alternative design
strategies are studied, namely, ($i$) increase the number of lobes
on each cam or ($ii$) increase the number of cams. This device is
intended to replace the current ball-screws in {\em Orthoglide}, a
three-DOF parallel robot for the production of translational
motions, currently under development at {\em Ecole Centrale de
Nantes} for machining applications.
\section{Introduction}
In robotics and mechatronics applications, whereby motion is
controlled using a piece of software, the conversion of motion
from rotational to translational is usually done by either {\em
ball screws} or {\em linear actuators}. Of these alternatives,
ball screws are gaining popularity, one of their drawbacks being
the high number of moving parts that they comprise, for their
functioning relies on a number of balls rolling on grooves
machined on a shaft; one more drawback of ball screws is their low
load-carrying capacity, stemming from the punctual form of contact
by means of which loads are transmitted. Linear bearings solve
these drawbacks to some extent, for they can be fabricated with
roller bearings; however, these devices rely on a form of
direct-drive motor, which makes them expensive to produce and to
maintain.

\begin{figure}[htb]
 \begin{minipage}[b]{8cm}
 \begin{center}
   \psfrag{Roller}{Roller}
   \psfrag{Follower}{Follower}
   \psfrag{Conjugate cams}{Conjugate cams}
   \centerline{\epsfig{file = 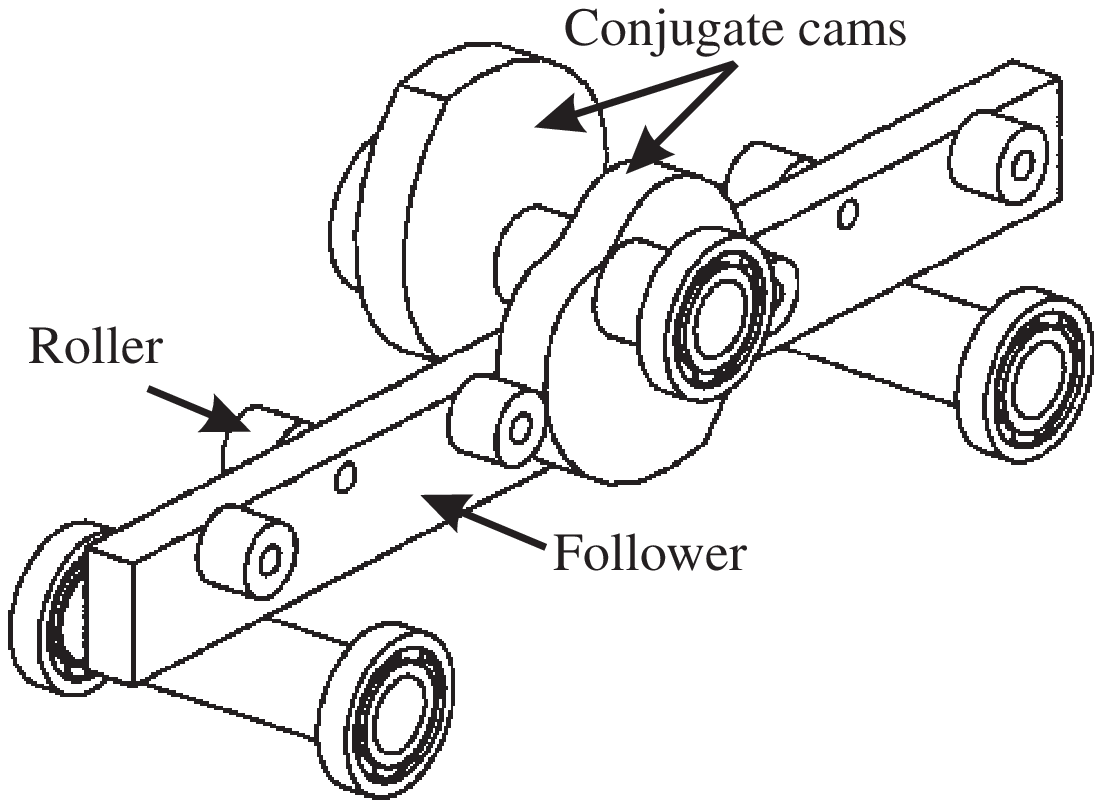,scale = 0.45}}
   \caption{Layout of \goodbreak Slide-o-Cam}
   \label{fig01}
 \end{center}
 \end{minipage}
 \begin{minipage}[b]{8cm}
 \begin{center}
    \centerline{\epsfig{file = 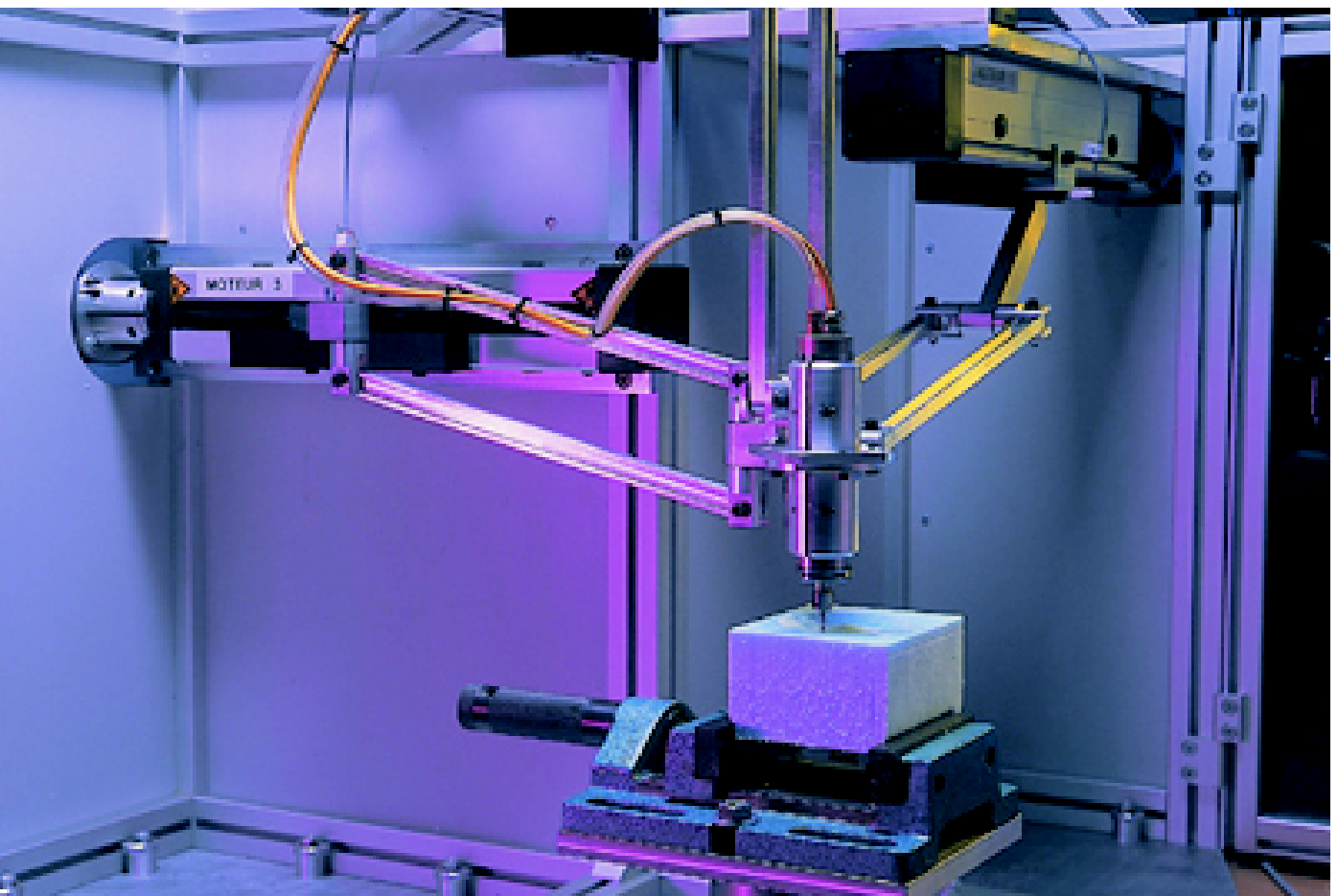,scale = 0.45}}
    \caption{The Orthoglide}
    \label{fig02}
 \end{center}
 \end{minipage}
 \end{figure}

A novel transmission, called {\em Slide-o-Cam}, was introduced in
\cite{Gonzalez-Palacios:2000} as depicted in Fig.~\ref{fig01}, to
transform a rotation into a translation. Slide-o-Cam is composed
of four major elements: ($i$) the frame, ($ii$) the cam, ($iii$)
the follower and ($iv$) the rollers. The input axis on which the
cam is mounted, the camshaft, is driven at a time-varying angular
velocity, by an actuator under computer-control. Power is
transmitted to the output, the translating follower, which is the
roller-carrying slider, by means of pure-rolling contact between
cam and roller. The roller, in turn, comprises two components, the
pin and the bearing. The bearing is mounted at one end of the pin,
while the other end is press-fit into the roller-carrying slider.
Contact between cam and roller thus takes place at the outer
surface of the bearing. The mechanism uses two conjugate
cam-follower pairs, which alternately take over the motion
transmission to ensure a positive action; the rollers are thus
driven by the cams throughout a complete cycle. The main advantage
of using a cam-follower mechanism instead of an alternative
transmission to transform rotation into translation is that
contact through a roller reduces friction, contact stress and
wear.

This transmission, once fully optimized, will replace the three
ball screws used by the Orthoglide prototype \cite{Chablat:2003}.
Orthoglide features three prismatic joints mounted orthogonaly,
three identical legs and a mobile platform, which moves in
Cartesian space with fixed orientation, as shown in
Fig.~\ref{fig02}. The three motors are SANYO DENKI (ref.
P30B08075D) with a constant torque of 1.2~Nm from 0 to 3000~rpm.
This property enables the mechanism to move throughout the
workspace a 4-kg load with an acceleration of 17~ms$^{-2}$ and a
velocity of 1.3~ms$^{-1}$. Furthermore, the pitch is 50~mm per cam
turn, while the minimum radius of the camshaft is 8.5~mm. A new
arrangement of camshaft, rollers and follower is proposed to
reduce the inertial load when more than two cams are needed.

Unlike Lampinen \cite{Lampinen:2003}, who used a genetic
algorithm, or Zhang and Shin \cite{Zhang:2004}, who used what they
called ``the relative-motion method,'' where the relative velocity and
acceleration of the the follower with respect to the cam is
prescribed, we use a deterministic method that takes into account
geometric and machining constraints \cite{Bouzakis:1997}.

The {\em pressure angle} $\mu$ (or its complement, the {\em
transmission angle}) is a key performance index of cam-follower
transmissions. One definition of {\em pressure angle} is {\em the
acute angle between the direction of the output force and the
direction of the velocity of the point where the output force is
applied} \cite{Shigley:1980}.
%

Moreover, unlike Carra, Garziera and
Pellegrini \cite{Carra:2004}, who used a negative radius-follower
to reduce the pressure angle, we use a positive radius-follower
that permits us to assemble several followers on the same
roller-carrier. To optimize the pressure angle, two alternative
design strategies are studied, namely, ($i$) increase the number
of lobes on each cam or ($ii$) increase the number of cams. The
relations defined in \cite{Renotte:2004} for cam profiles with one
lobe are extended to several lobes.
\section{Synthesis of the Planar Cam Mechanism}
\begin{figure}[htb]
 \begin{minipage}[b]{8cm}
 \begin{center}
    \psfrag{O1}{$O_1$}
    \psfrag{O2}{$O_2$}
    \psfrag{mu}{$\mu$}
    \psfrag{C}{$C$}
    \psfrag{f}{$\negr f$}
    \psfrag{x}{$x$}
    \psfrag{y}{$y$}
    \psfrag{e}{$e$}
    \psfrag{s}{$s$}
    \psfrag{u}{$u$}
    \psfrag{v}{$v$}
    \psfrag{a4}{$a_4$}
    \psfrag{p_n}{$p/n$}
    \psfrag{P}{$\psi$}
    \epsfig{file = 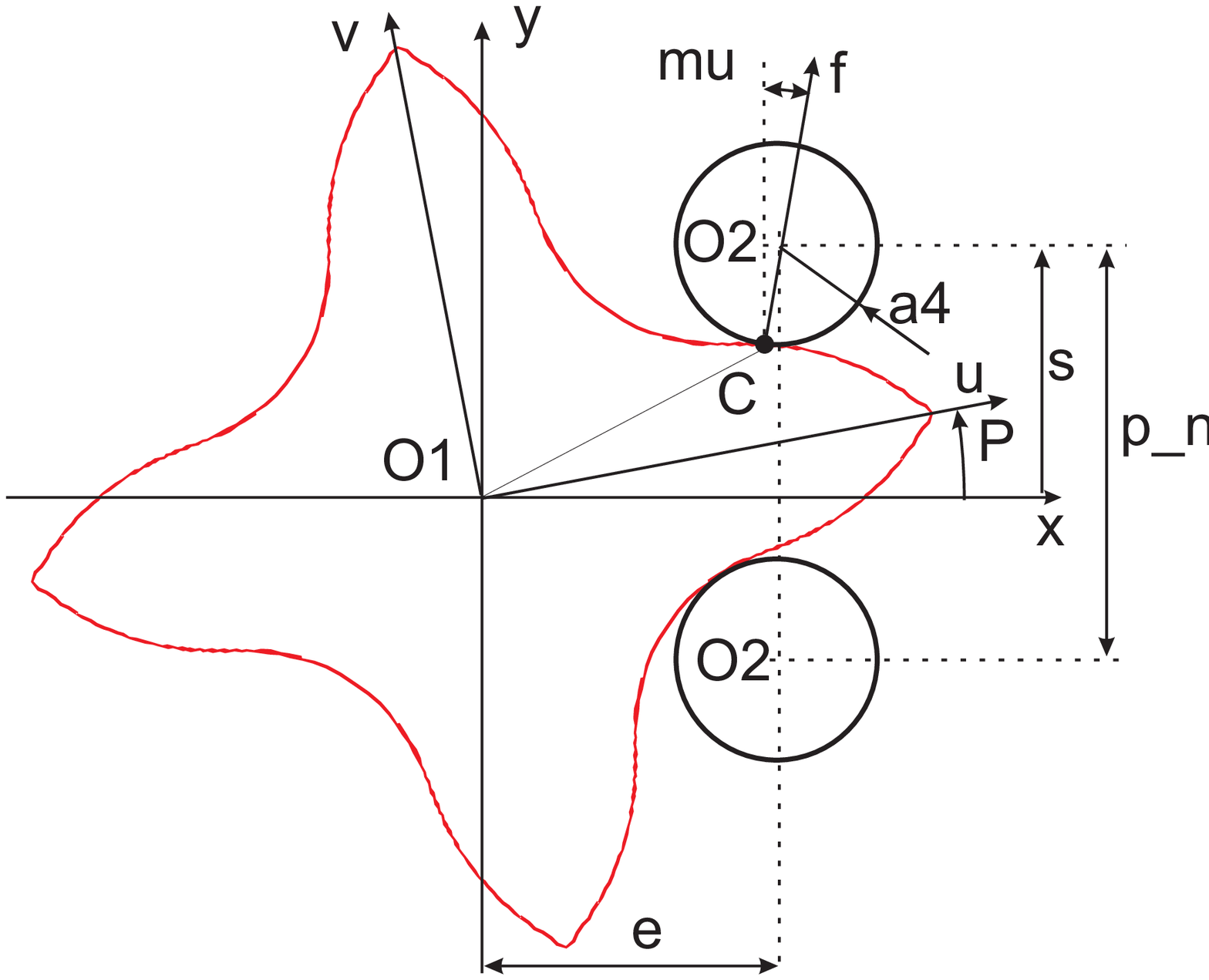,scale = 0.3}
    \caption{Parameterization of Slide-O-Cam}
    \label{fig04}
 \end{center}
 \end{minipage}
 \begin{minipage}[b]{8cm}
 \begin{center}
    \psfrag{O1}{$O_1$}
    \psfrag{O2}{$O_2$}
    \psfrag{C}{$C$}
    \psfrag{x}{$x$}
    \psfrag{y}{$y$}
    \psfrag{S(0)}{$s(0)$}
    \psfrag{u}{$u$}
    \psfrag{v}{$v$}
    \psfrag{p_n}{$p/n$}
    \epsfig{file = 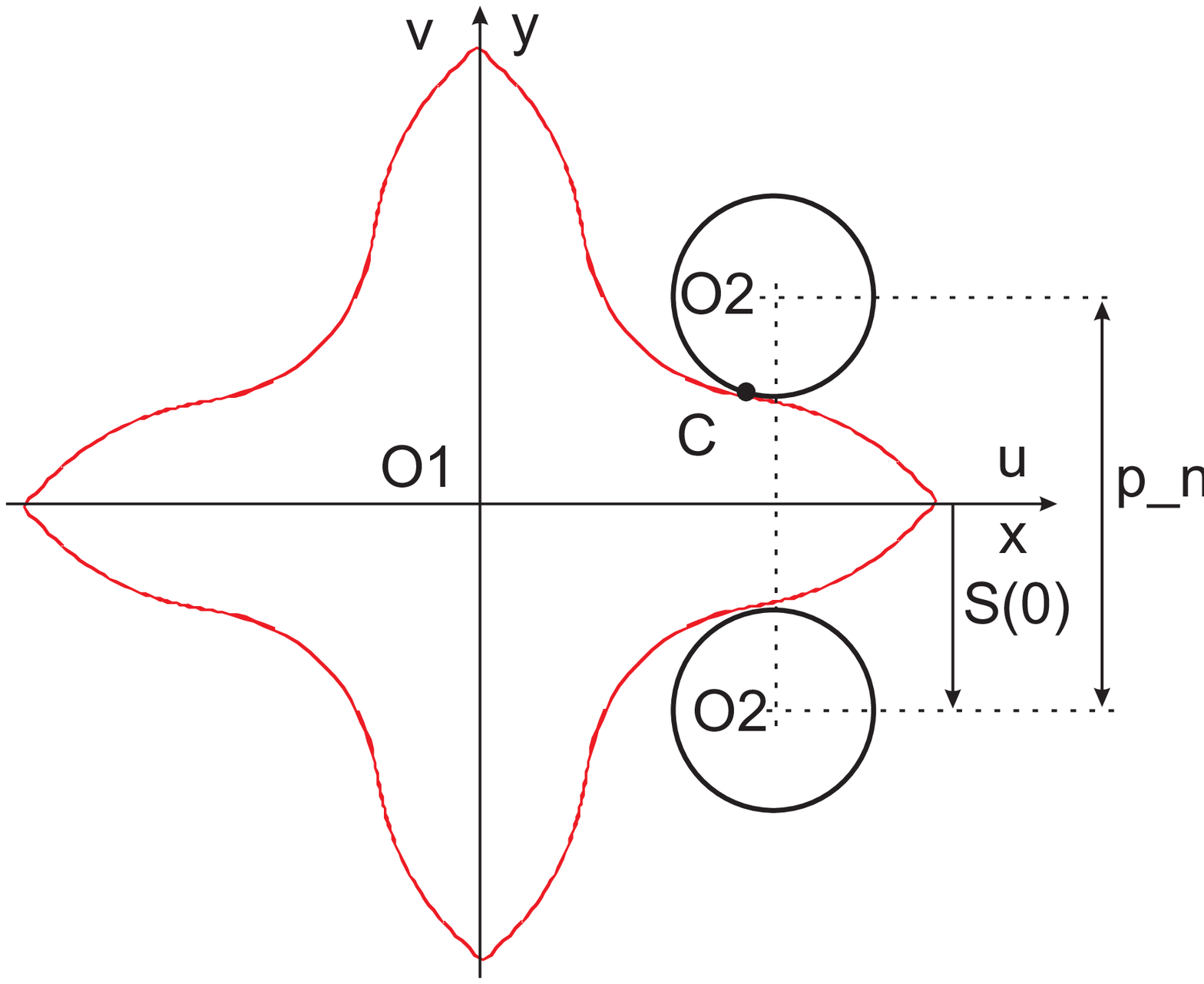,scale = 0.3}
    \caption{Home configuration of the mechanism, for a cam with four lobes}
    \label{fig05}
 \end{center}
 \end{minipage}
 \end{figure}

Let the $x$-$y$ frame be fixed to the machine frame and the
$u$-$v$ frame be attached to the cam, as depicted in
Fig.~\ref{fig04}. $O_{\!1}$ is the origin of both frames, while
$O_{2}$ is the center of the roller, and $C$ is the contact point
between cam and roller. The geometric parameters defining the cam
mechanism, with $n$ lobes, are illustrated in the same figure. The
notation of this figure is based on the general notation
introduced in \cite{Gonzalez-Palacios:1993} and complemented in
\cite{Renotte:2004}, namely,
\begin{itemize}
\item $p$, the pitch, {\em i.e.}, the distance between the center of two
rollers on the same side of the follower, in single-lobed cams (distance is
$p/n$ in $n$-lobbed cams);
\item $n$, the number of lobes;
\item $e$, the distance between the axis of the cam and the line of centers of
the rollers;
\item $a_{4}$, the radius of the roller bearing, {\em i.e.}, the radius of the
roller;
\item $\psi$, the angle of rotation of the cam, the input of the mechanism;
\item $s$, the displacement of the follower, given by the the position of the
center of the roller, namely, the output function of the mechanism;
\item $\mu$, the pressure angle;
\item $\bf f$, the force transmitted from cam to roller.
\end{itemize}
In our design, $p$ is set to 50~mm, in order to meet the Orthoglide
specifications.
\subsection{The Input-Output Function}
The above parameters as well as the contact surface on the cam are
determined by the geometric relations dictated by the
Aronhold-Kennedy Theorem in the plane \cite{Waldron:1999}. When
the cam makes a complete turn ($\Delta \psi=2\pi$), the
displacement of the roller is equal to $p$, {\em i.e.}, the
distance between $n+1$ rollers on the same side of the
roller-carrying slider ($\Delta s =p$). Furthermore, if we
consider the home configuration of the roller, as depicted in
Fig.~\ref{fig05}, the roller is below the $x$-axis for $\psi=0$,
so that we have $s(0)=-p/(2n)$. Hence, the expression for the
input-output function $s$ is
 \be
   s(\psi)=\frac{p}{2\pi}\psi-\frac{p}{2n} \label{eq01}
 \ee
The expressions for the first and second derivatives of $s(\psi)$
with respect to $\psi$ will be needed; these are readily derived,
namely,
 \be
  s'(\psi)=\frac{p}{2\pi} \quad {\rm and} \quad s''(\psi)=0
 \label{eq02}
 \ee
\subsection{Cam-Profile Determination}
The cam profile is determined by the displacement of the contact
point $C$ around the cam. This contact between one lobe and one
roller takes place within $0 \leq \psi \leq 2\pi/n$. For this
domain we find the profile of one lobe. The remaining $n-1$ lobes
are found by rotation around $O_1$. The Cartesian coordinates of
$C$ in the $u$-$v$ frame take the form
\cite{Gonzalez-Palacios:1993}
 \begin{eqnarray}
 u_{c}(\psi)\!\!\!\! &=&
 \! b_{2} \cos(\psi)+(b_{3}-a_{4})\cos(\delta-\psi) \nonumber \\
 v_{c}(\psi)\!\!\!\! &=&
 \!\!\!\! -b_{2} \sin(\psi)+(b_{3}-a_{4})\sin(\delta-\psi)\nonumber
 \end{eqnarray}
\newline with coefficients $b_{2}$, $b_{3}$ and $\delta$ given by
 \begin{subequations}
 \begin{eqnarray}
  b_{2}  \!\!\!\!&=&\!\! -s'(\psi) \sin \alpha_{1} \\
  b_{3}  \!\!\!\!&=&\!\! \sqrt{[e+s'(\psi)  \sin \alpha_{1}]^{2}+[s(\psi)
\sin \alpha_{1}]^{2}} \\
  \delta \!\!\!\!&=&\!\!  \arctan \left( \frac{-s(\psi) \sin \alpha_{1}}{e+s'(\psi) \sin
 \alpha_{1}} \right)
 \end{eqnarray}
 \label{eq04}
 \end{subequations}
\newline where $\alpha_{1}$ is the directed angle between the axis of the
cam and the translating direction of the follower; $\alpha_{1}$ is positive in the ccw direction. Considering the sign-convention adopted for the input angle $\psi$ and for the output $s$, as depicted in Fig.~\ref{fig04}, we have
 \begin{equation}
  \alpha_{1}=-\pi /2 \label{eq05}
 \end{equation}
We introduce now the non-dimensional design parameter $\eta$,
which will be extensively used:
 \begin{equation}
  \eta=\frac{e}{p} \label{eq06}
 \end{equation}
Thus, from Eqs.(\ref{eq01}), (\ref{eq02}a), (\ref{eq04}a--c),
(\ref{eq05}) and (\ref{eq06}), we derive the expressions for
coefficients $b_{2}$, $b_{3}$ and $\delta$:
 \begin{subequations}
  \begin{eqnarray}
  b_{2}  \!\!\!&=&\!\!\!  \frac{p}{2\pi} \\
  b_{3}  \!\!\!&=&\!\!\!  \frac{p}{2\pi}\sqrt{(2\pi\eta-1)^{2}+
\left(\psi-\frac{\pi}{n}\right)^{2}} \\
  \delta \!\!\!&=&\!\!\! \arctan \left(\frac{n\psi-\pi}{2n\pi\eta-n}\right)
  \end{eqnarray}
  \label{eq07}
 \end{subequations}
\newline whence a first constraint on $\eta$, $\eta \neq 1/(2\pi)$, is
derived. With such a definition, however, the cam profile does not
close.

\begin{figure}[ht]
 \begin{center}
     \psfrag{O1}{$O_1$}
     \psfrag{x}{$x$}
     \psfrag{y}{$y$}
     \psfrag{u}{$u$}
     \psfrag{v}{$v$}
     \psfrag{C}{$C$}
     \psfrag{Psi}{$\psi$}
     \psfrag{-u}{-$u$}
     \psfrag{-v}{-$v$}
     \psfrag{(a)}{(a)}
     \psfrag{(b)}{(b)}
     \psfrag{(c)}{(c)}
     \psfrag{(d)}{(d)}
     \epsfig{file =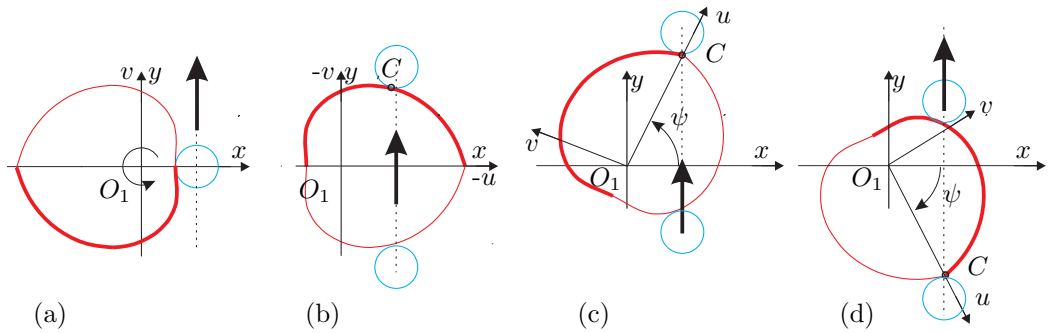,scale = 0.7}
 \end{center}
 \caption{Extended angle $\Delta$ for different values of $\psi$, for a cam
with one lobe: (a) $\psi=\pi$; (b) $\psi=2\pi$; (c)
$\psi=2\pi-\Delta$; and (d) $\psi=\Delta$}
 \label{fig06}
 \end{figure}

An {\em extended angle} $\Delta$ is introduced \cite{Lee:2001}, so
that the cam profile closes. Angle $\Delta$ is obtained as the
root of the equation $v_{c}(\psi)=0$. In the case of Slide-o-Cam,
$\Delta$ is negative, as shown in Fig.~\ref{fig06} for a cam with
one lob. The contact between the cam profile and one roller is
made (resp. is lost) for $\psi=\Delta$ or $\psi=2\pi-\Delta$ as
depicted in Fig.~\ref{fig06} (c\&d).

Consequently, the cam profile closes within $\Delta \leq \psi \leq
2\pi/n-\Delta$, with a corresponding relation for the remaining
$n-1$ lobes.
\subsection{Pitch-curve Determination}
The pitch curve is the trajectory of the center $O_{2}$ of the
roller, distinct from the trajectory of the contact point $C$,
which produces the cam profile. The Cartesian coordinates of
points $O_{2}$ in the $x$-$y$ frame are $(e,s)$, as depicted in
Fig.~\ref{fig04}. Hence, the Cartesian coordinates of the pitch
curve in the $u$-$v$ frame are
 \begin{subequations}
 \begin{eqnarray}
  u_{p}(\psi) \!\!\!& = &\!\! ~e\cos (\psi)+s(\psi)\sin (\psi)  \\
  v_{p}(\psi) \!\!\!& = &\!\! -e\sin (\psi)+s(\psi)\cos (\psi)
 \end{eqnarray}
 \label{eq09}
 \end{subequations}
 \begin{figure}
 \begin{center}
     \psfrag{O1}{$O_1$}
     \psfrag{x}{$x$}
     \psfrag{y}{$y$}
     \psfrag{(A)}{(a)}
     \psfrag{(B)}{(b)}
     \psfrag{(C)}{(c)}
     \psfrag{(D)}{(c)}
     \psfrag{-15}{-15}\psfrag{-10}{-10}\psfrag{-5}{-5}
     \psfrag{15} {15} \psfrag{10} {10} \psfrag{5} {5}
     \epsfig{file =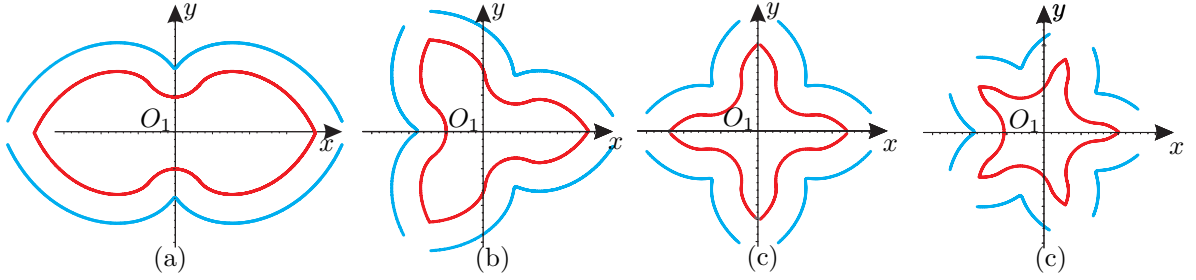,scale = 1}
     \caption{Cam profile (in red) and pitch curve (in blue), within $\Delta \leq \psi \leq 2\pi/n-\Delta$, with $a_4=4$, $p=50$ and $e=9$, for:
     (a) $n=2$; (b) $n=3$; (c) $n=4$; and (d) $n=5$.}
     \label{fig07}
 \end{center}
 \end{figure}
Figure~\ref{fig07} shows a plot of the cam profile and its pitch
curve, within $\Delta \leq \psi \leq 2\pi/n-\Delta$, throughout
all $n$ lobes.
\subsection{Geometric Constraints on the Mechanism}
In order to lead to a feasible mechanism, the radius $a_{4}$ of
the roller must satisfy two conditions, as shown in
Fig.~\ref{fig08}:
\begin{figure}
\center
  \psfrag{O1}{$O_1$}
  \psfrag{O2}{$O_2$}
  \psfrag{p_n}{$p/n$}
  \psfrag{x}{$x$}
  \psfrag{y}{$y$}
  \psfrag{p}{$p$}
  \psfrag{e}{$e$}
  \psfrag{b}{$b$}
  \psfrag{C}{$C$}
  \psfrag{B}{$B$}
  \psfrag{a4}{$a_4$}
  \epsfig{file =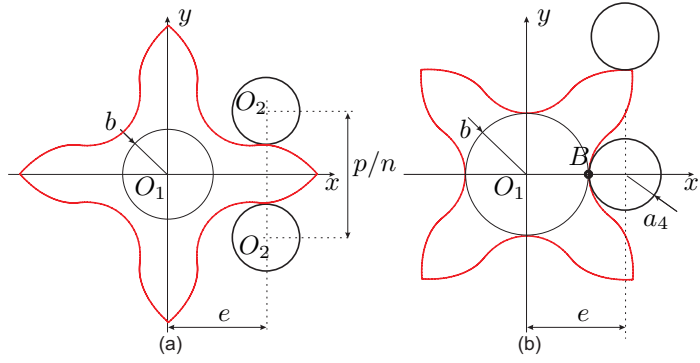,scale = 0.3}
  \caption{Constraints on the radius of the roller: (a) $ a_{4} < p/(2n)$; and (b) $a_{4} \leq \eta p-b$.}
  \label{fig08}
\end{figure}
\begin{itemize}
\item
Two consecutive rollers on the same side of the roller-carrying
slider must not interfere with each other. Since $p/n$ is the
distance between the center of two consecutive rollers, we have
the constraint $2a_{4} < p/n$. Hence the first condition on
$a_{4}$:
     \begin{equation}
     \label{eq010}
     a_{4} < \frac{p}{2n}
     \end{equation}
\item
The radius $b$ of the shaft on which the cams are mounted must be
taken into consideration. Hence, we have the constraint $a_{4}+b
\leq e $ , the second constraint on $a_{4}$ in terms of the
parameter $\eta$ thus being\footnote{ It is possible to have
$a_{4} = \eta p-b$ if the cams and the shaft are machined from the
same block. Thus, there is a common point, referred to as $B$ in
Fig.~\ref{fig08}b, between the shaft and the cam profile. This
will not be the solution chosen for our design, because it is too
complicated to machine the cams and the shaft on the same block.}
     \begin{equation}
     \label{eq011}
     a_{4} \leq \eta p-b
     \end{equation}
\end{itemize}
Considering the initial configuration of the roller, as depicted
in Fig.~\ref{fig05}, the $v$-component of the Cartesian coordinate
of the contact point $C$ is negative in this configuration, {\em
i.e.}, $v_{c}(0) \leq 0$. However, from the expression for
$v_{c}(\psi)$ and for parameters $b_{3}$ and $\delta$ given in
Eqs.~(\ref{eq07}b~\&~c), respectively, the above relation leads to
the condition:
 \be
 \left( \frac{p}{2\pi n}\sqrt{(2n\pi\eta-n)^{2}+\pi^{2}} - a_{4} \right)
 \sin\left[\arctan\left(\frac{-\pi}{2n\pi\eta -n}\right) \right] \leq 0
 \ee
Further, we define parameters $A$ and $B$ as
 \be
 A=\frac{p}{2n\pi}\sqrt{(2n\pi\eta-n)^{2}+\pi^{2}} - a_{4}
 \quad \mbox{and} \quad
 B=\sin \left[\arctan\left(\frac{-\pi}{2n\pi\eta -n}\right) \right]
 \ee
Now we derive a constraint $A$. Since $(2\pi n\eta-n)^{2}
>0$, we have
 \beqa
 \sqrt{(2n\pi\eta-n)^{2}+\pi^{2}} > \pi \nonumber
 \eeqa
Hence,
 \beqa
    A >\frac{p}{2n}-a_{4} \nonumber
 \eeqa
Furthermore, from the constraint on $a_{4}$ established in
Eq.~(\ref{eq010}), we have $p/(2n)-a_{4} > 0$, whence $A>0$.
Consequently, the constraint $v_{c}(0) \leq 0$ leads to the
constraint $B \leq 0$.

\begin{figure}[hb]
\center
  \psfrag{0.15}{$1/(2\pi)$}
  \psfrag{0.30}{$1/\pi$}
  \psfrag{0.45}{$0.45$}
  \psfrag{1}{$1$}
  \psfrag{2}{$2$}
  \psfrag{3}{$3$}
  \psfrag{4}{$4$}
  \psfrag{A4}{$a_4$}
  \psfrag{mu}{$\eta$}
  \psfrag{Eta}{$\eta$}
  \psfrag{N}{$n$}
  \psfrag{feasible domain}{feasible domain}
  \epsfig{file =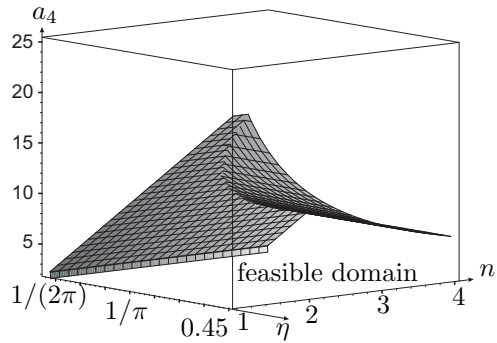,scale = 0.3}
  \caption{Design parameter space for $P=50mm$ and $b=4.25mm$}
  \label{fig15}
\end{figure}

We transform now the expression for $B$ by using the trigonometric
relation:
 \beqa
 \forall x \in \mathbb{R},~ \sin(\arctan
 x)=\frac{x}{\sqrt{1+x^{2}}} \nonumber
 \eeqa
Hence, the constraint $v_{c}(0) \leq 0$ becomes
 \[
 \frac{-\pi}{n(2\pi\eta-1) \sqrt{1+\pi^{2}/(2n\pi\eta-n)^{2}}} \leq 0
 \]
which holds only if $2\pi \eta-1 > 0$. Finally, the constraint
$v_{c}(0) \leq 0$ leads to a constraint on $\eta$,
namely\footnote{If in the initial configuration the roller were on
the upper side of the $x$-axis, the input-output function would be
$s(\psi)=(p/2\pi)\psi+p/2$, and we would have the constraint
$v_{c}(0) \geq 0$, which would lead to the same constraint
$\eta>1/(2\pi)$.},
 \begin{equation}
  \label{eq012}
  \eta>\frac{1}{2\pi}
 \end{equation}
Equations~\ref{eq010}, \ref{eq011} and \ref{eq012} permit us to
reduce the design parameter space. Figure~\ref{fig15} depicts the
design parameter space for $P=50mm$ and $b=4.25mm$. We can notice
that when the number of lobes by cam increases, the maximum value
of $a_4$ decreases.
\subsection{Pressure Angle}
The pressure angle of cam-roller-follower mechanisms is defined as
the angle between the common normal at the cam-roller contact
point $C$ and the velocity of $C$ as a point of the follower
\cite{Angeles:1991}, as depicted in Fig.~\ref{fig04}, where the
pressure angle is denoted by $\mu$. This angle plays an important
role in cam design. The smaller $|\mu|$, the better the force
transmission. In the case of high-speed operations, {\em i.e.},
angular velocities of cams exceeding 50~rpm, the pressure-angle is
recommended to lie within 30$^{\circ}$.

For the case at hand, the expression for the pressure-angle $\mu$
is given in \cite{Angeles:1991} as
 \[
    \tan\mu=\frac{s'(\psi)-e}{s(\psi)}
 \]
Considering the expressions for $s$ and $s'$, and using the
parameter $\eta$ given in Eqs.(\ref{eq01}), (\ref{eq02}a) and
(\ref{eq07}), respectively, the expression for the pressure angle
becomes
 \[
  \tan\mu=\frac{n-2n\pi\eta}{n\psi-\pi}
 \]
\subsection{Conjugate Cams}
\begin{figure}[ht]
  \begin{center}
  \psfrag{O1}{$O_1$}
  \psfrag{x}{$x$}
  \psfrag{y}{$y$}
  \psfrag{-5}{-$5$}
  \psfrag{-10}{-$10$}
  \psfrag{-15}{-$15$}
  \psfrag{-20}{-$20$}
  \psfrag{-30}{-$30$}
  \psfrag{5}{$5$}
  \psfrag{10}{$10$}
  \psfrag{15}{$15$}
  \psfrag{20}{$20$}
  \psfrag{30}{$30$}
  \psfrag{(a)}{(a)}
  \psfrag{(b)}{(b)}
  \epsfig{file =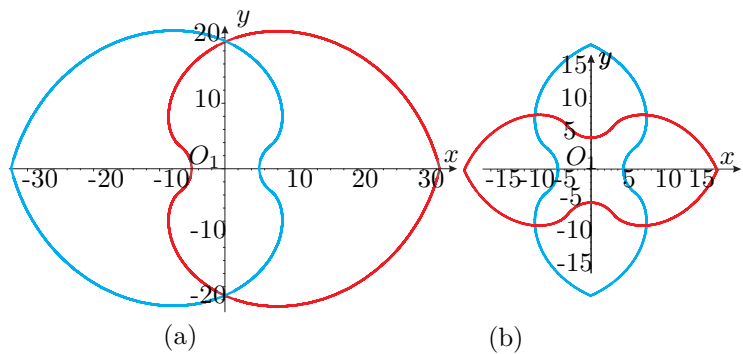,scale = 0.45}
  \caption{Assembly of two cams with $p=50$, $a_4=10$ and $e=9$: (a) one lobe; (b) two lobes}
  \label{fig09}
  \end{center}
\end{figure}

To reduce the pressure angle, several cams can be assembled in the
same cam-shaft. We note $m$ the number of cams and $\beta$ the
angle of rotation between two adjacent cams, {\em i.e.},
 \beqa
    \beta= \frac{2 \pi}{nm} \nonumber
 \eeqa
On the Slide-o-Cam mechanism first designed in
\cite{Gonzalez-Palacios:2000}, two conjugate cams with one lobe
each and $\beta=\pi$ were used. Figure~\ref{fig09} shows two cam
profiles with one and two lobes.
\subsection{Convexity of the Cam Profile}
The convexity of the cam influences the accuracy of the machining;
for this reason it is preferable to have a convex cam profile for
each part of the cam where it drives the roller. In this section,
we establish conditions on the design parameters $\eta$ and
$a_{4}$ in order to have such a property. We will thus study the
sign of the curvature of the cam profile {\em via} that of the
pitch curve.

The curvature of any planar parametric curve, in terms of the
Cartesian coordinates $u$ and $v$, and parameterized with any
parameter $\psi$, is given in \cite{Angeles:1991},
 \begin{equation}
 \label{eq17}
 \kappa=\frac{v'(\psi)u''(\psi)-u'(\psi)v''(\psi)}{[u'(\psi)^{2}+v'(\psi)^{2}]^{3/2}}
 \end{equation}
The sign of $\kappa$ in Eq.(\ref{eq17}) tells whether the curve is
convex or concave at a point: a positive $\kappa$ implies a
convexity, while a negative $\kappa$ implies a concavity at that
point.
The computation of the first and second derivatives of the
Cartesian coordinates of the pitch curve, defined in
Eqs.~(\ref{eq09}), with respect to the angle of rotation of the cam,
$\psi$, are
 \begin{subequations}
 \begin{eqnarray}
 u'_{p}(\psi)  \!\!& = & ~~(~s'(\psi)-e)\sin(\psi)+s(\psi)\cos(\psi)  \\
 v'_{p}(\psi)  \!\!& = & ~~(~s'(\psi)-e)\cos(\psi)-s(\psi)\sin(\psi)  \\
 u''_{p}(\psi) \!\!& = & ~~(2s'(\psi)-e)\cos(\psi)-s(\psi)\sin(\psi)  \\
 v''_{p}(\psi) \!\!& = & -(2s'(\psi)-e)\sin(\psi)-s(\psi)\cos(\psi)
 \end{eqnarray}
\label{eq18}
\end{subequations}
By substituting $\eta=e/p$, with Eqs.(\ref{eq18}a-d) into
Eq.(\ref{eq17}), the curvature $\kappa_{p}$ of the pitch curve can
be obtained as
\begin{equation}
\label{eq19}
\kappa_{p}=-\frac{2n^{2}\pi}{p}\frac{[(n\psi-\pi)^{2}+2n^{2}(2\pi\eta-1)(\pi\eta-1)]}{[(n\psi-\pi)^{2}+n^{2}(2\pi\eta-1)^{2}]^{3/2}}
\end{equation}
provided that the denominator never vanishes for any value of
$\psi$, {\em i.e.}, if we observe the condition:
 \[
    \eta \neq 1/(2\pi)
 \]
Considering the expression for $\kappa_{p}$ in Eq.(\ref{eq19}), we
have, for every value of $\psi$,
 \[
 \kappa_{p} \geq 0 ~~~\mbox{if}~~~ (2\pi \eta-1)(\pi \eta-1) \geq 0 ~~\mbox{and}~~ \eta \neq \frac{1}{2\pi}
 \]
whence the condition on $\eta$:
 \begin{equation}
  \label{eq21}
  \kappa_{p} \geq 0 ~~~\mbox{if}~~~
  \eta \in [0,\frac{1}{2\pi}[ \cup [\frac{1}{\pi}, + \infty[
 \end{equation}
The condition on $\eta$ given in Eq.(\ref{eq21}) must be combined
with the condition established in Eq.(\ref{eq012}), $\eta >
1/(2\pi)$. Hence, the final \textbf{convexity condition of the
pitch curve} is:
 \be
  \label{eq22} \eta \geq \frac{1}{\pi}
 \ee
\section{Influence of the Number of Conjugate Cams and the Number of Lobes on the
Pressure Angle}
\subsection{Active Angular Interval of the Pressure Angle}
\begin{figure}[hc]
  \begin{center}
  \psfrag{mu}{$\mu$ (degree)}
  \psfrag{psi}{$\psi$(rad)}
  \psfrag{-pi}{$-\pi$}
  \psfrag{pi}{$\pi$}
  \psfrag{2pi}{$2\pi$}
  \psfrag{12}{12} \psfrag{10}{10}   \psfrag{8}{8} \psfrag{7}{7}
  \psfrag{6}{6}   \psfrag{5}{5}     \psfrag{4}{4} \psfrag{3}{3}
  \psfrag{2}{2}   \psfrag{0}{0}
  \psfrag{-2}{-2} \psfrag{-4}{-4}   \psfrag{-6}{-6}
  \psfrag{-20}{-20}  \psfrag{-40}{-40}  \psfrag{-60}{-60}  \psfrag{-80}{-80}
  \psfrag{20}{20}    \psfrag{40}{40}    \psfrag{60}{60}    \psfrag{80}{80}
  \psfrag{t_c_p}{the came pushes}
  \psfrag{t_t_l}{to the left}
  \psfrag{t_t_r}{to the right}
  \psfrag{pi-d}{$\pi-\Delta$}
  \psfrag{2pi-d}{$2\pi-\Delta$}
  \epsfig{file =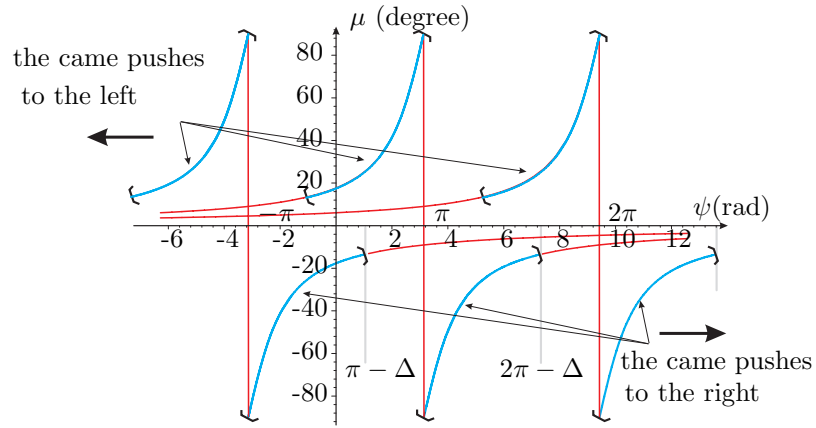,scale = 0.40}
  \caption{Pressure-angle distribution for one cam with one lobe, with $p=50$, $a_4=10$ and $e=9$}
  \label{fig10}
  \end{center}
\end{figure}

\begin{figure}[ht]
  \begin{center}
  \psfrag{mu}{$\mu$ (degree)}
  \psfrag{psi}{$\psi$(rad)}
  \psfrag{12}{12} \psfrag{10}{10}   \psfrag{8}{8} \psfrag{7}{7}
  \psfrag{6}{6}   \psfrag{5}{5}     \psfrag{4}{4} \psfrag{3}{3}
  \psfrag{2}{2}   \psfrag{0}{0}
  \psfrag{-2}{-2} \psfrag{-4}{-4}   \psfrag{-6}{-6}
  \psfrag{-20}{-20}  \psfrag{-40}{-40}  \psfrag{-60}{-60}  \psfrag{-80}{-80}
  \psfrag{20}{20}    \psfrag{40}{40}    \psfrag{60}{60}    \psfrag{80}{80}
  \psfrag{t_c_p}{the came pushes}
  \psfrag{t_t_l}{to the left}
  \psfrag{t_t_r}{to the right}
  \psfrag{(A)}{(a)}
  \psfrag{(B)}{(b)}
  \epsfig{file =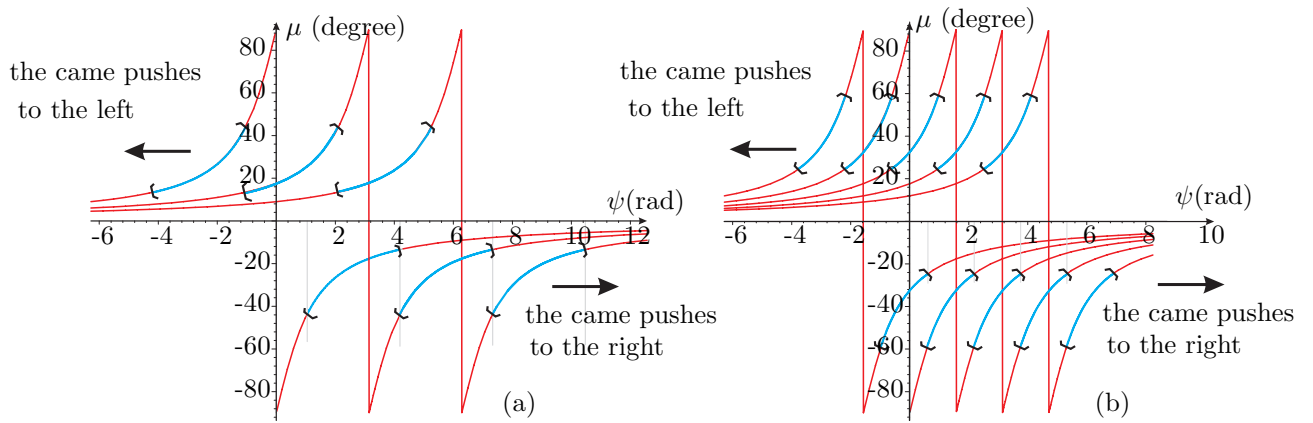,scale = 0.40}
  \caption{Pressure-angle distribution for two conjugate-cams with (a) one lobe and (b) two lobes, with $p=50$, $a_4=10$ and $e=9$}
  \label{fig11}
  \end{center}
\end{figure}

We are only interested in the interval of $\psi$ where the cam
drives the roller to go to the right; we call this the {\em active
interval}. An other interval of $\psi$ exists, with the same
range, when the cam drives the roller to go to the left as is
depicted in Fig.~\ref{fig10}.
\begin{itemize}
\item {\bf A single-cam mechanism:}
The active interval is:
\end{itemize}
 \[
 \frac{\pi}{n} \leq \psi \leq \frac{2\pi}{n}-\Delta
 \]
Indeed, if we start the motion in the home configuration depicted
in Fig.~\ref{fig05}, with the cam rotating in the ccw direction,
the cam begins to drive the roller only when $\psi=\pi/n$; the cam
can drive the follower until contact is lost, {\em i.e.}, when
$\psi=2\pi/n-\Delta$, as shown in Fig.~\ref{fig06}c. With only one
cam, the cam cannot drive the roller throughout one complete turn,
as depicted in Fig.~\ref{fig10}. This result holds for all values
of $n$. Thus, a cam-follower assembly must have a least two
conjugate cams. When $-\pi/n+\delta \leq \psi \leq 2\pi/n-\delta$,
contact between two followers and the cam occurs.
\begin{itemize}
\goodbreak
\item {\bf A two-conjugate-cam mechanism:}
The active interval is:
\end{itemize}
 \[
 \frac{\pi}{n}-\Delta \leq \psi \leq
 \frac{2\pi}{n}-\Delta
 \]
Indeed, the conjugate cam can also drive the follower when $0 \leq
\psi \leq \pi/n-\Delta$; there is, therefore, a common interval,
for $\pi/n \leq \psi \leq \pi/n-\Delta$, during which two cams can
drive the follower. In this interval, the conjugate cam can drive
a roller with lower absolute values of the pressure angle. We
assume that, when the two cams drive the rollers, the cam with the
lower absolute value of pressure angle effectively drives the
follower. Consequently, we are only interested in the value of the
pressure angle in the interval $\pi/n-\Delta \leq \psi \leq
2\pi/n-\Delta$, as depicted in Fig.~\ref{fig11}.

\begin{figure}
  \begin{center}
  \psfrag{mu}{$\mu$ (degree)}
  \psfrag{psi}{$\psi$(rad)}
  \psfrag{12}{12} \psfrag{10}{10}   \psfrag{8}{8} \psfrag{7}{7}
  \psfrag{6}{6}   \psfrag{5}{5}     \psfrag{4}{4} \psfrag{3}{3}
  \psfrag{2}{2}   \psfrag{0}{0}
  \psfrag{-2}{-2} \psfrag{-4}{-4}   \psfrag{-6}{-6} \psfrag{-5}{-5}
  \psfrag{-20}{-20}  \psfrag{-40}{-40}  \psfrag{-60}{-60}  \psfrag{-80}{-80}
  \psfrag{20}{20}    \psfrag{40}{40}    \psfrag{60}{60}    \psfrag{80}{80}
  \psfrag{t_c_p}{the came pushes}
  \psfrag{t_t_l}{to the left}
  \psfrag{t_t_r}{to the right}
  \epsfig{file =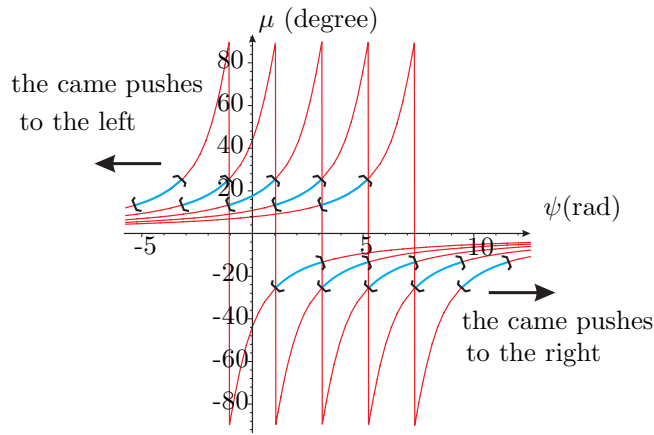,scale = 0.40}
  \caption{Pressure-angle distribution for three conjugate-cams with one lobe, $p=50$, $a_4=10$ and $e=9$}
  \label{fig12}
  \end{center}
\end{figure}

\begin{itemize}
\item {\bf A three-conjugate-cam mechanism:}
For the same reason, the active interval is:
\end{itemize}
 \[
 \frac{4\pi}{3n}-\Delta \leq \psi \leq
 \frac{2\pi}{n}-\Delta
 \]
This interval is 33\% smaller than a two-conjuage-cam-mechanism,
the part of the interval removed having the highest pressure
angle, as depicted in Fig.~\ref{fig12}.
\subsection{Pressure Angle and Design Parameters}
\begin{figure}
 \begin{center}
  \psfrag{mu}{$\mu$ (degree)}
  \psfrag{psi}{$\psi$(rad)}
  \psfrag{eta=5}{$\eta=5$}
  \psfrag{eta=2}{$\eta=2$}
  \psfrag{eta=1.5}{$\eta=1.5$}
  \psfrag{eta=1}{$\eta=1$}
  \psfrag{eta=0.8}{$\eta=0.8$}
  \psfrag{eta=0.4}{$\eta=0.4$}
  \psfrag{eta=1/pi}{$\eta=1/\pi$}
  \psfrag{8}{8} \psfrag{7}{7} \psfrag{6}{6} \psfrag{5}{5} \psfrag{4}{4} \psfrag{3}{3} \psfrag{2}{2} \psfrag{1}{1}\psfrag{0}{0}
  \psfrag{-20}{-20}  \psfrag{-40}{-40}  \psfrag{-60}{-60}  \psfrag{-80}{-80}
  \psfrag{(a)}{(a)}
  \psfrag{(b)}{(b)}
  \psfrag{(c)}{(c)}
  \psfrag{(d)}{(d)}
  \epsfig{file =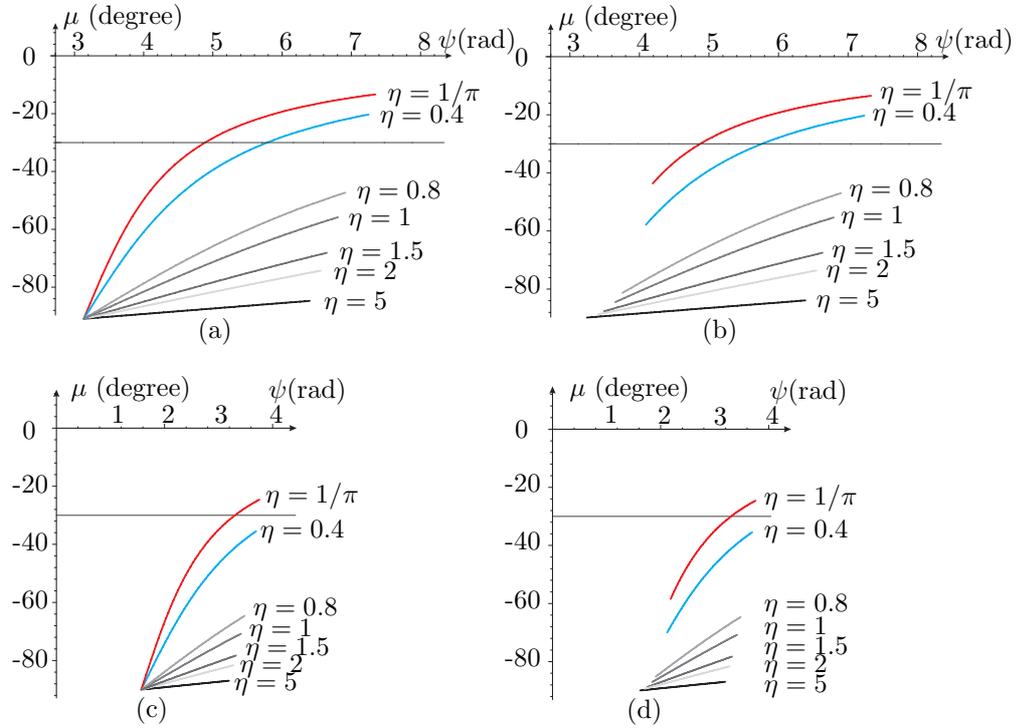,scale = 0.35}
  \caption{Influence of parameter $\eta$ on the pressure angle $\mu$
(in degree), with $p=50$~mm and $a_{4}=10$~mm, for one or two
lobes and one or two cams, for (a) one cam and one lobe; (b) two
cams and one lobe; (c) one cam and two lobes; and (d) two cams and
two lobes}
  \label{fig13}
  \end{center}
 \end{figure}

\begin{figure}
 \begin{center}
  \psfrag{mu}{$\mu$ (degree)}
  \psfrag{psi}{$\psi$(rad)}
  \psfrag{eta=5}{$\eta=5$}
  \psfrag{eta=2}{$\eta=2$}
  \psfrag{eta=1.5}{$\eta=1.5$}
  \psfrag{eta=1}{$\eta=1$}
  \psfrag{eta=0.8}{$\eta=0.8$}
  \psfrag{eta=0.4}{$\eta=0.4$}
  \psfrag{eta=1/pi}{$\eta=1/\pi$}
  \psfrag{8}{8} \psfrag{7}{7} \psfrag{6}{6} \psfrag{5}{5} \psfrag{4}{4} \psfrag{3}{3} \psfrag{2}{2} \psfrag{1}{1}\psfrag{0}{0}
  \psfrag{-20}{-20}  \psfrag{-40}{-40}  \psfrag{-60}{-60}  \psfrag{-80}{-80}
  \psfrag{(a)}{(a)}
  \psfrag{(b)}{(b)}
  \psfrag{(c)}{(c)}
  \psfrag{(d)}{(d)}
  \epsfig{file =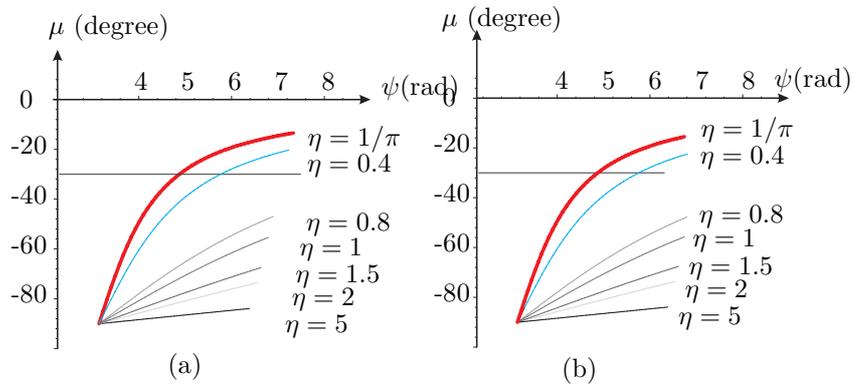,scale = 0.6}
  \caption{Influence of parameter $a_4$ on the pressure angle $\mu$
(in degree), for one lobe and two cams, with $p=50$~mm and (a)
$a_{4}=10$~mm and (b) $a_4=25~mm$}
  \label{fig14}
  \end{center}
 \end{figure}

We study here the influence of design parameters $\eta$, $a_{4}$
and $n$ on the values of the pressure angle, whith the cam driving
the roller, as reported in \cite{Zhang:2004b}. We also study the
influence of the number of conjugate cams on the value of pressure
angle.
 \begin{itemize}
 \item{\bf Influence of parameter $\eta$}
 \end{itemize}
Figure~\ref{fig13} shows the influence of the parameter $\eta$ on
the pressure angle, with $a_{4}$ and $p$ being fixed for several
numbers of cams and lobes. From these plots we have the result:
 \begin{quote}
 \textit{The lower $\eta$, the lower the absolute value of the
pressure angle, with $\eta \geq 1/\pi$, as defined in
Eq.\ref{eq22}.}
 \end{quote}
This result is the same if we consider a single cam with several
lobes or a two- or three-conjugate-cam mechanism.
 \begin{itemize}
\item {\bf Influence of the radius of the roller $a_{4}$}
 \end{itemize}
Parameter $a_{4}$ does not appear in the expression for the
pressure angle, but it influences the value of the extended angle
$\Delta$, and hence the plot boundaries of the pressure angle, as
shown in Fig.~\ref{fig14} for two value of $a_4$.

By computing the value of the extended angle $\Delta$ for several
values of $a_{4}$, we notice that the higher $a_{4}$, the lower
the absolute value of $\Delta$. Consequently, since the boundaries
to plot the pressure angle for a system with two conjugate cams
are $\pi/n-\Delta$ and $2\pi/n-\Delta$ or $4\pi/3n-\Delta$ and
$2\pi/n-\Delta$ for three conjugate cams, we notice that when we
increase $a_{4}$, $-\Delta$ decreases or, equivalently, $\Delta$
increases and the boundaries are translated toward the left, {\em
i.e.}, toward higher absolute values of the pressure angle. We
thus have the result:
\begin{quote}
\textit{The lower $a_{4}$, the lower the absolute value of the
pressure angle.}
\end{quote}
This result is valid for single or conjugate-cam mechanisms and is
independent of the number of lobes.
\begin{itemize}
\item {\bf Influence of the number of lobes $n$}
\end{itemize}
By computing the value of the pressure angle for several values of
$n$, as depicted in Fig.~\ref{fig13}, we have the result:
\begin{quote}
\textit{The lower $n$, the lower the absolute value of the
pressure angle.}
\end{quote}
For at least two lobes, contact is lost when the pressure angle is
greater than $20^{\circ}$.
\begin{itemize}
\item {\bf Influence of the Number of Conjugate Cams}
\end{itemize}
Figures~\ref{fig11}(a) and \ref{fig12} show that the higher the
number of cams, the lower the active interval. Especially, the
number of cams reduces the maximum absolute value of the pressure
angle. Therefore,
\begin{quote}
\textit{The higher the number of conjugate cams, the lower the
absolute value of the pressure angle.}
\end{quote}
\section{Conclusions}
Two design strategies have been used in this paper to reduce the
pressure angle in the Slide-o-Cam mechanism. The optimum solution
is found whenever ($i$) the number of lobes is one, ($ii$) for the
higher number of conjugate cams, ($iii$) for the smallest value of
the radius of the roller $a_4$ and ($iv$) for the smallest value
of $\eta$. However, to minimize the deformation of the roller pin,
$a_4$ have to be computed to resist to efforts during the motion
as is made in \cite{Renotte:2004}. Further research is currently
underway to evaluate the influence of variations in the design
parameters on the pressure angle and the active interval and the
design of cam with non-convex shape.
\bibliographystyle{unsrt}

\end{document}